\documentclass[10pt,twocolumn,letterpaper]{article}

\usepackage[pagenumbers]{cvpr} %

\usepackage{graphicx}
\usepackage{amsmath}
\usepackage{amssymb}
\usepackage{booktabs}
\usepackage{float}
\usepackage{colortbl}
\usepackage{enumitem}

\providecommand{\tabularnewline}{\\}
\definecolor{red}{rgb}{0.97,0.79,0.68}
\definecolor{blue}{rgb}{0.7,0.77,0.90}
\definecolor{navy}{rgb}{0.0,0.0,0.9}

\usepackage[pagebackref,breaklinks,colorlinks,allcolors=navy]{hyperref}
\usepackage[T1]{fontenc}
\usepackage[utf8]{inputenc}

\usepackage[capitalize]{cleveref}
\crefname{section}{Sec.}{Secs.}
\Crefname{section}{Section}{Sections}
\Crefname{table}{Table}{Tables}
\crefname{table}{Tab.}{Tabs.}

\def\cvprPaperID{9256} %
\def\confName{CVPR}
\def\confYear{2022}

\begin{document}

\title{Disentangling visual and written concepts in CLIP}

\author{Joanna Materzy\'nska\\
MIT\\
{\tt\small jomat@mit.edu}
\and
Antonio Torralba\\
MIT\\
{\tt\small torralba@mit.edu}
\and
David Bau\\
Harvard\\
{\tt\small davidbau@seas.harvard.edu}
}

\twocolumn[{%
\vspace{-1em}
\maketitle
\vspace{-1em}

\begin{center}
    \centering 
    \vspace{0.15in}
    \captionsetup{type=figure}
    \includegraphics[width=1\linewidth]{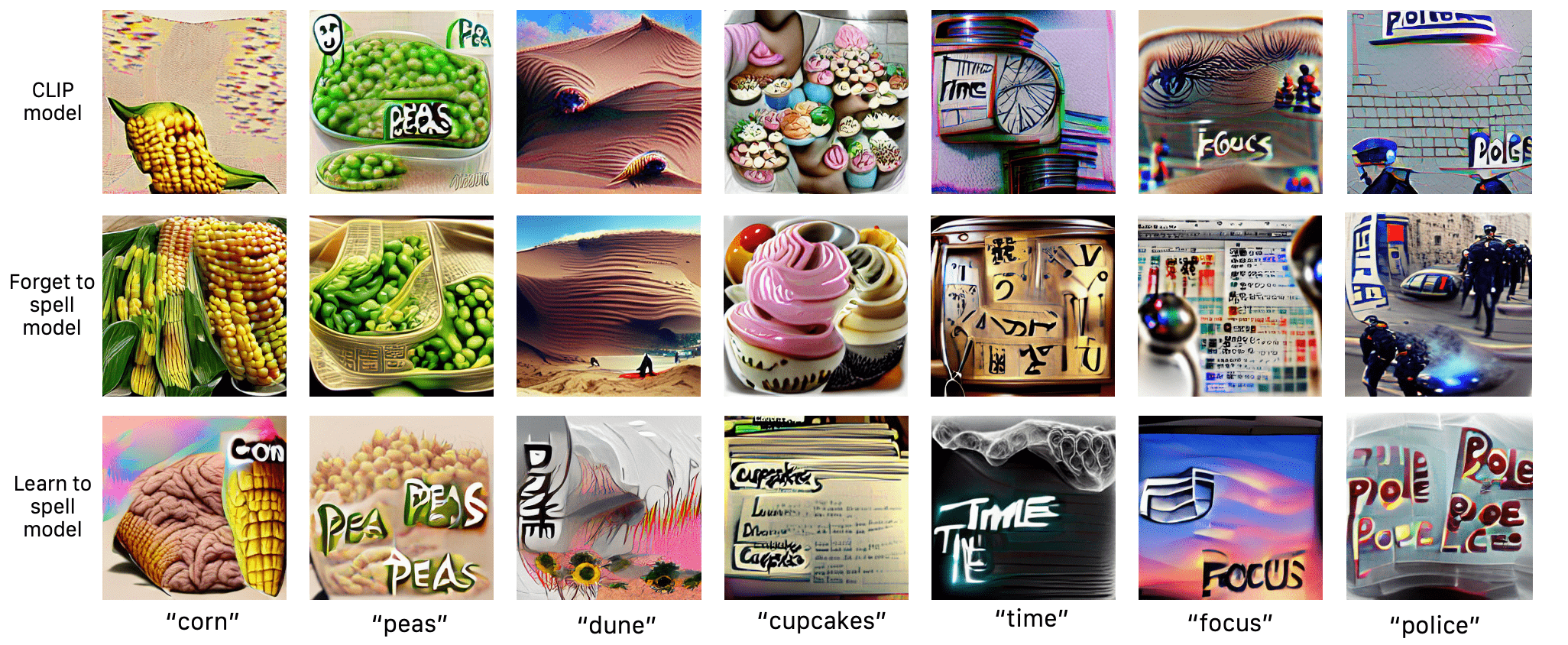}
    \vspace{-0.1in}
    \captionof{figure}{Generated images conditioned on text prompts (top row) disclose the entanglement of written words and their visual concepts. Our proposed orthogonal projections of the vector space disentangle the space into one corresponding to visual concepts (middle row), and written words (bottom row).}
    \label{fig:bigteaser}
\end{center}
}]

\begin{abstract}

\vspace{0.1in}
The CLIP network measures the similarity between natural text and images; in this work, we investigate the entanglement of the representation of word images and natural images in its image encoder.  First, we find that the image encoder has an ability to match word images with natural images of scenes described by those words. This is consistent with previous research that suggests that the meaning and the spelling of a word might be entangled deep within the network. On the other hand, we also find that CLIP has a strong ability to match nonsense words, suggesting that processing of letters is separated from processing of their meaning. To explicitly determine whether the spelling capability of CLIP is separable, we devise a procedure for identifying representation subspaces that selectively isolate or eliminate spelling capabilities.  We benchmark our methods against a range of retrieval tasks, and we also test them by measuring the appearance of text in CLIP-guided generated images.  We find that our methods are able to cleanly separate spelling capabilities of CLIP from the visual processing of natural images.,\footnote{The project website, source code and dataset are available at  \href{https://joaanna.github.io/disentangling_spelling_in_clip/}{ https://joaanna.github.io/disentangling\_spelling\_in\_clip/}.} 

\end{abstract}

\section{Introduction}
\label{sec:intro}
\begin{figure}[t]
  \centering
  \includegraphics[width=\linewidth]{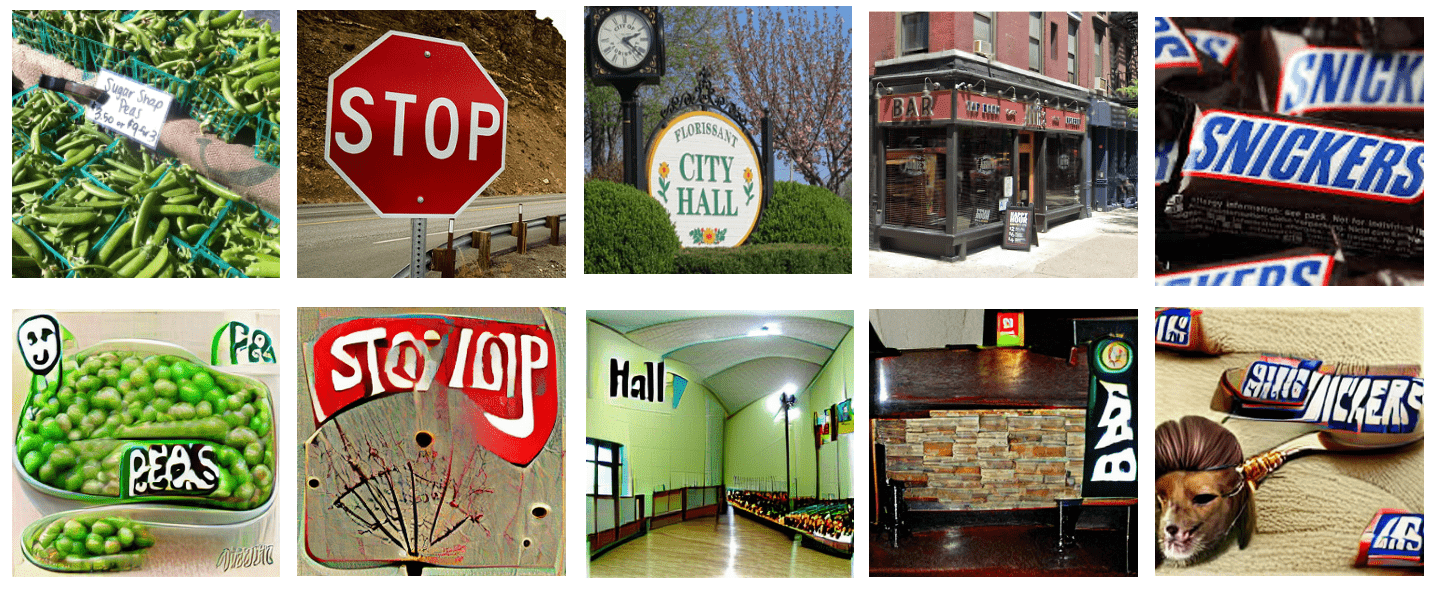}
   \caption{Top row: examples of written text in natural images, bottom row: generated images conditioned on words ("peas", "stop sign", "hall", "bar", "snickers").}
   \label{fig:motivation}
\end{figure}

The distinction between written words and visual objects is crystal clear for us: we would never confuse an object with a written word describing that object.  However, it has been shown \cite{goh2021multimodal} that attaching a white sheet of paper with “iPad” written on it to an apple, will cause a neural network to shift its prediction to lean towards what is written instead of recognizing the fruit.  We hypothesize that the network learns to confuse text with objects because of the prevalence of text in real-world training data: text on products, signs, and labels is often visible next to the thing it represents (Figure~\ref{fig:motivation}), which is perhaps why a neural network would struggle to distinguish an object from its written name. Beginning with a pretrained network that exhibits this text/object confusion, we ask if the perception of text by a network can be separated from the perception of objects.

We study the representations of the CLIP \cite{radford2021learning} network, which is trained to measure the similarity between natural text and images, and which has been shown to be vulnerable to confusion between written text and visual concepts \cite{goh2021multimodal,lemesle2022language}. In \cite{goh2021multimodal}, feature visualizations of neurons within CLIP revealed the presence of “multi-modal neurons” that activate when presented with different forms of the same concept; for example, the same neuron will activate on an image of a written word and an image of the object described by that word. In addition to this, we have found that text-to-image generation methods that use CLIP will spell out the word they have been conditioned on (Figure~\ref{fig:bigteaser}). Together, these findings indicate a deeply rooted correlation between written words and their visual concepts in the image encoder of CLIP.

In this paper, we investigate how CLIP makes sense of written words, and whether CLIP distinguishes its understanding of written words from their visual meaning. Specifically, %
we investigate whether the image encoding permits separation of information about written words from the visual concepts described by those words. We find that a simple setup and an orthogonal projection can in fact separate the two capabilities. We demonstrate applications of this disentanglement by removing text artifacts in text-to-image generation, and by defending against typographic attacks. We collect a dataset of 180 images of 20 objects and 8 attacks and measure the confusion between the true object labels and typographic attacks between the CLIP model and our disentangled representation. We find that in both distinct applications, the effect of text is greatly reduced.

\section{Related Works}

\textbf{Understanding Representations} Our work follows the tradition of a line of approaches for understanding the internal representations of a model by training a small model on the representation:  \cite{alain2016understanding}  proposed training simple classifier probes for testing the presence of information in a network; \cite{zhou2018interpretable} observes that such linear probes can be used to create explanations of a decision and \cite{fong2018net2vec} uses such probing models to map a dictionary of concepts through a network. Conversely, \cite{kim2018interpretability} proposes using gradients of a simple classifier  to estimate the sensitivity of a network to a classified concept, and to distinguish between causal and correlative effects. Our work to identify the text processing subspace within CLIP differs from previous methods because we use a contrastive loss to identify a large representation subspace for information about visual words. Rather than measuring classification accuracy, we verify our findings by applying the probed model to generate images.  Concurrent work \cite{lemesle2022language} applies cognitive science tools and finds evidence that the vision and language do not share semantic representation in CLIP network, consistent with our findings.

\textbf{Controllable GAN Generation}
Increasingly powerful image GAN models have sparked interest in steerable image generation methods that synthesize an image by guiding the generator towards some objective:   GAN output can be steered by directly guiding generation towards target images \cite{jahanian2020steerability}; or by optimizing loss of a classifier \cite{goetschalckx2019ganalyze,shen2020interfacegan}; or PCA, clustering or other methods can also be used to directly identify meaningful representation subspaces for manipulating a GAN \cite{harkonen2020ganspace,collins2020editing,wu2021stylespace}.  The release of CLIP~\cite{radford2021learning}, a large-scale model to score text-and-image similarity has unleashed a wave of creativity, because it enables any generative model to be guided by open text. The state-of-the-art  DALL-E~\cite{ramesh2021zero} uses CLIP; and CLIP has also been combined with StyleGAN~\cite{karras2020analyzing,patashnik2021styleclip,bau2021paint}, BigGAN~\cite{murdock2021bigsleep}, and VQGAN~\cite{esser2021taming,crowson2021vqganclip, crowsonvqgan}.  Like these methods, we investigate the ability of CLIP to steer VQGAN, however instead of generating individual images, we ask whether the broad ability of CLIP to read and draw visual words can be controlled.

\section{Terminology}
To avoid confusion while discussing words within images, we begin by defining some terminology.

\vspace{1ex}
\small
\noindent
Kinds of images:
\begin{itemize}[leftmargin=*]
    \item \textbf{image text}:
    \begin{itemize}
        \item \textbf{synthetic image text} : an image of text rendered on a white background
        \item \textbf{image text in the wild}: text on a signboard found in a photograph of a real scene
    \end{itemize}
    \item \textbf{natural images}: images depicting the real world
    \item \textbf{natural image with text}: natural image is modified by adding rendered text
    \item \textbf{natural image with word class label}: natural image with text, where the text is a class name
\end{itemize}
Kinds of text: 
\begin{itemize}[leftmargin=*]
    \item \textbf{ text class label}: the text name of a class  category, composed by prepending a string “an image of a” to the name
 \item \textbf{text string}: a word as processed by a text encoder; this could be either a real English word or a fake nonsense string, composed of random letters
\end{itemize}
\normalsize

\begin{figure}
  \centering
  \includegraphics[width=\linewidth]{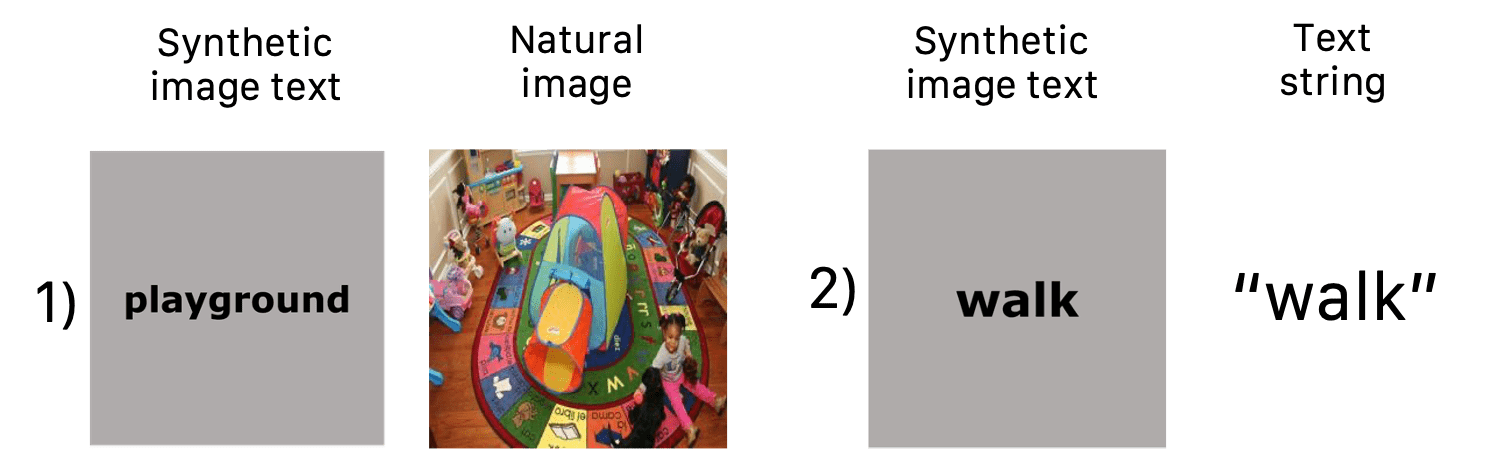}
   \caption{Visual comprehension tasks, 1) associating natural images with word class label images, 2) word image and language word retrieval.}
   \label{fig:visual_tasks}
\end{figure}

\vspace{-0.05in}
\section{Visual comprehension}

Does the image encoder of CLIP encode image text differently from the way it encodes the visual concept described by that same text?

We investigate this question by measuring the ability of CLIP to solve a task that it was not originally trained to do: rather than matching natural images with text strings as encoded by the text encoder, we test the ability of CLIP to match natural images with image text as encoded by the CLIP image encoder, discarding the text encoder entirely.  For example, we ask whether the CLIP image encoder will match visual image text of the word ``playground'' with a natural image of a playground scene. (Figure~\ref{fig:visual_tasks})

We consider two datasets, Places 365~\cite{zhou2017places} and ImageNet~\cite{russakovsky2015imagenet}, and report the top-1 validation accuracy of our task in Table~\ref{tab:image_reading}. This visual comprehension task achieves 15.58\% top-1 accuracy on Places 365 and 10.58\% top-1 accuracy on ImageNet. While accuracy is lower than zero-shot image-to-text classification, our result is far better than random, and it confirms our hypothesis that the CLIP image encoder correlates written words with their visual meaning.%
\begin{table}
    \centering
          \resizebox{\linewidth}{!}{
\begin{tabular}{c|c|c}
Model & \multicolumn{2}{c}{Top-1 Accuracy}\tabularnewline
\hline 
 & Places 365 & ImageNet\tabularnewline
\hline 
CLIP ViT-B/32 ZS with PE & 39.47 &63.36 \tabularnewline
\hline
CLIP ViT-B/32 ZS without PE & 37.25 &56.72 \tabularnewline
\hline 
CLIP image to image class & 15.58  &  10.58\tabularnewline
\hline 
Random baseline & 0.1 & 0.27 \tabularnewline
\end{tabular}}
\caption{Image classification as visual comprehension task, ZS denotes zero-shot and PE prompt engineering.}
    \label{tab:image_reading}
\end{table}%

Next we investigate if CLIP relies on understanding the meaning of a word to read a word.  In particular, we ask how well CLIP can associate \textit{any} string, including both real English words, and fake word nonsense strings, created by uniformly sampling letters from the Latin alphabet of length ranging from 3 to 8.  We form image text with these strings, and we compute the retrieval score (1 out of 20k) on the set of real, fake and all strings and report the results in Table~\ref{tab:text_reading}. Strikingly, we observe that CLIP is able to retrieve both real words and nonsense strings, despite (most likely) never having seen those nonsense strings in natural images. 

This leads us to the question: how does the image encoder of CLIP read?  Is its reading capability separated from its other visual processing, for example as a distinct capability to recognize and spell out individual letters?  Or is its OCR deeply entangled with its understanding of real words, inseparable from the perception of natural images described by that word?  To resolve that question, we design and benchmark a method to disentangle text and natural image processing.%

\begin{table}[t]
  \centering
  \resizebox{\linewidth}{!}{
\begin{tabular}{c|c|c|c}
{} & {{\small \# image text and text string}} & \multicolumn{2}{c}{Retrieval score}\tabularnewline
 &  & Img2Txt & Txt2Img\tabularnewline
\hline 
All strings & 40 000 & 60.66 & 75.97\tabularnewline
\hline 
Real words & 20 000 & 76.38 & 91.46\tabularnewline
\hline 
Nonsense strings & 20 000 & 61.77 & 79.19\tabularnewline
\end{tabular}}
\caption{Text to image retrieval on real words and nonsense strings.}
    \label{tab:text_reading}
\end{table}

\section{Disentangling Text and Vision with Linear Projections}

Motivated by the deeply rooted confusion between written text and visual concepts, we aim to disentangle the CLIP vector space's visual space from the written one. Our approach is to identify an orthogonal, lower-dimensional projection of the learned representations to achieve this goal. To this end, we collect a dataset consisting of tuples with five elements $(x_i, y_i, x_t, y_t, x_{it})$. The first two elements $(x_i, y_i)$ are natural images and their text class labels. Image texts and text strings $(x_t, y_t)$, and $x_{it}$ being the natural image $x_i$ with the string from the synthetic image text $x_t$ rendered on it.

\begin{figure}[!ht]
  \centering
  \includegraphics[width=0.9\linewidth]{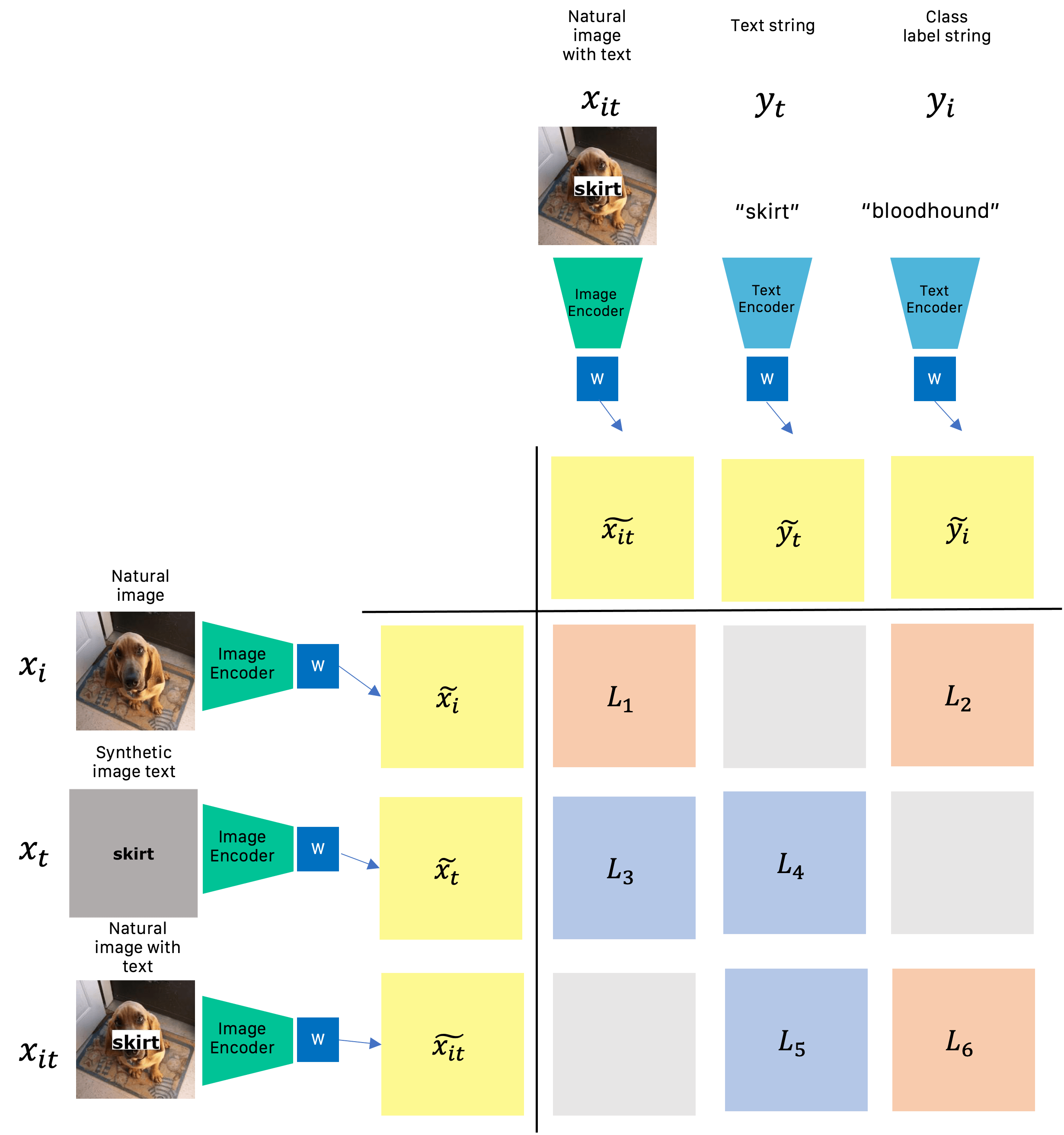}
   \caption{In our method, different pairs from the tuple  $(x_i, y_i, x_t, y_t, x_{it})$ are trained to minimize their distance in the projection space. The losses in red correspond to the task of visual concepts, and the losses in blue to the distilling written words.}
   \label{fig:method}
\end{figure}

We precompute the CLIP embeddings of the images and text prompts using CLIP vision and text encoders, and train an orthogonal matrix $W$ for each of the tasks. During training, depending on the task, we apply a symmetric cross entropy $L_i$ on the given pair of embeddings, following the CLIP training procedure. We also introduce a regularizer term to the loss $\mathcal{R}(W) = \| \mathbb{I} - WW^T\|$ that encourages W to be orthogonal.

We call the projection that captures the written concepts in the network: ``learn to spell'' model. This model should be able to respond well to the text and images of text hence, the embeddings of the image texts $x_t$ and the embedding of the text strings $y_t$ should be close in space, similarly a natural image with text $x_{it}$ should be close to either the image text and text strings ($x_t, y_t$). Those losses are shown in blue in Figure~\ref{fig:method}. The losses shown in red correspond to the opposite task, learning to ignore the written text in natural images. Thus, during training the ``learn to spell'' model, we maximize the red objectives and minimize the blue objectives. The overall loss can be written as:
\small
\begin{equation}
    L_{spell} = - L_1 - L_2 - L_6 +  L_3 +L_4 + L_5 +\gamma \mathcal{R}(W)
\end{equation}
\label{eq:1}
\normalsize
The ``forget to spell'' model, that focuses on the visual parts in images, will conversely aim to minimize the red and maximize the blue objectives.
\small%
\begin{equation}
    L_{forget} =  L_1 + L_2 + L_6 -  L_3 - L_4 -  L_5 +\gamma \mathcal{R}(W)
\end{equation}
    \label{eq:2}
\normalsize
We empirically test the effects of the contributing loss terms and present results in section 6.1.

\begin{figure}
  \centering
  \includegraphics[width=.85\linewidth]{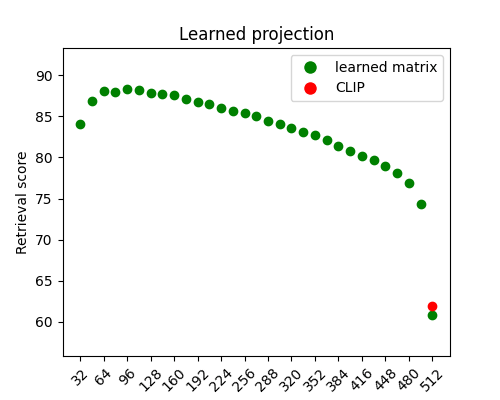}
   \caption{Varying bottleneck dimension of the learned projection matrix versus retrieval score on the text retrieval task.}
   \label{fig:bottleneck}
\end{figure}
\section{Experiments}

\begin{table*}[ht]
  \centering
  \resizebox{\textwidth}{!}{
\begin{tabular}{c|c|c|c|c|c|c|c|c|c|c|c|c|c|c|c|c}
\multicolumn{6}{c}{} &  & \multicolumn{2}{c|}{Top-1 Accuracy} & \multicolumn{8}{c}{Retrieval Accuracy {[}img2txt{]}}\tabularnewline
\cline{8-17} \cline{9-17} \cline{10-17} \cline{11-17} \cline{12-17} \cline{13-17} \cline{14-17} \cline{15-17} \cline{16-17} \cline{17-17} 
\multicolumn{6}{c}{Loss} &  & {\cellcolor{red} $\downarrow$ ($x_i$, $y_i$)} & {\cellcolor{red} $\downarrow$ ($x_{it}$, $y_i$)} & \multicolumn{2}{c|}{\cellcolor{blue}$\uparrow$ ($x_{t}$, $y_t$)} & \multicolumn{2}{c|}{\cellcolor{blue} $\uparrow$ ($x_{it}$, $x_t$)} & \multicolumn{2}{c|}{\cellcolor{red} $\downarrow$ ($x_{it}$, $x_i$)} & \multicolumn{2}{c}{ \cellcolor{blue} $\uparrow$($x_{it}$, $y_t$)}\tabularnewline
\cline{10-17} \cline{11-17} \cline{12-17} \cline{13-17} \cline{14-17} \cline{15-17} \cline{16-17} \cline{17-17} 
\multicolumn{6}{c}{} &  &  &  & {real} & {fake} & {real} & {fake} & {real} & {fake} & {real} & {fake}\tabularnewline
\cline{1-7} \cline{2-7} \cline{3-7} \cline{4-7} \cline{5-7} \cline{6-7} \cline{7-7} 
\cellcolor{red} $L_1$ & \cellcolor{red} $L_2$ &\cellcolor{blue} $L_3$& \cellcolor{blue} $L_4$ & \cellcolor{blue} $L_5$ & \cellcolor{red} $L_6$ & $\mathcal{R}(W)$ &  &  &  &  &  &  &  &  &  & \tabularnewline
\hline 
 &  &  &  &  &  &  & 56.72 & 33.04 & 76.27 & 61.88 & 98.87 & 95.64 & 89.97 & 89.53 & 62.57 & 48.52\tabularnewline
\hline
\hline
  &  &  & \checkmark &  &  & 0.5 & 0.99 & 0.16 & 89.62 & 87.58 & 99.00 & 98.13 & 4.29  & 2.36 & 84.01 & 79.69\tabularnewline
\hline 
  &  &  &  \checkmark& \checkmark &   & 0.5 & 0.52 & 0.12 & 90.88 & 87.59 & 99.46 & 98.93 & 1.29 & 1.06 & 88.81 & 83.81\tabularnewline
\hline 
 \checkmark &  &   & \checkmark & \checkmark &   & 0.5 & 0.2 & 0.13 & 90.86 & 87.49 & 99.43 & 98.94 & 1.19 & 0.94 & 88.58 & 83.96\tabularnewline
\hline 
  &  & \checkmark &  \checkmark & \checkmark &  & 0.5 & 0.51 & 0.11 & 91.86 & 88.06 & 99.54 & 99.06 & 1.22 & 1.05 & 90.28 & 84.75\tabularnewline
\hline 
 \checkmark &  &  \checkmark &  \checkmark & \checkmark &   & 0.5 & 0.19 & 0.13 & 91.89 & 88.15 & 99.55 & 99.1 & 1.21 & 0.98 & 90.3 & 84.77\tabularnewline
\hline 
 \checkmark &  &  \checkmark &  \checkmark &  \checkmark &  \checkmark & 0.5 & 0.17 & 0.06 & 89.81 & 87.49 & 99.29 & 99.00 & 1.22 & 1.02 & 87.43 & 83.51\tabularnewline
\hline 
 \checkmark &  \checkmark &  \checkmark &  \checkmark &  \checkmark &  \checkmark & 0.5 & 0.01 & 0.05 & 84.11 & 85.0 & 99.25 & 98.9 & 1.56 & 1.06 & 81.13 & 80.32\tabularnewline
\hline 
\hline
 \checkmark &  &  \checkmark &  \checkmark & \checkmark &   & 0.5 & 0.19 & 0.13 & 91.89 & 88.15 & 99.55 & 99.1 & 1.21 & 0.98 & 90.3 & 84.77\tabularnewline
\hline 
\checkmark &  & \checkmark & \checkmark & \checkmark &  & 0.0 & 0.08 & 0.08 & 82.07 & 79.86 & 98.19 & 97.88 & 0.6 & 0.23 & 76.78 & 74.38\tabularnewline
\end{tabular}
}
\caption{The ablation of the effects of different loss terms across classification and retrieval tasks of the tuples on the validation set for the "learn to spell" model.}
    \label{tab:retireval_learn}
\end{table*}
\begin{table*}[ht]
  \centering
  \resizebox{\textwidth}{!}{
\begin{tabular}{c|c|c|c|c|c|c|c|c|c|c|c|c|c|c|c|c}
\multicolumn{6}{c}{} &  & \multicolumn{2}{c|}{Top-1 Accuracy} & \multicolumn{8}{c}{Retrieval Accuracy {[}img2txt{]}}\tabularnewline
\cline{8-17} \cline{9-17} \cline{10-17} \cline{11-17} \cline{12-17} \cline{13-17} \cline{14-17} \cline{15-17} \cline{16-17} \cline{17-17} 
\multicolumn{6}{c}{Loss} &  & {\cellcolor{red} $\downarrow$ ($x_i$, $y_i$)} & {\cellcolor{red} $\downarrow$ ($x_{it}$, $y_i$)} & \multicolumn{2}{c|}{\cellcolor{blue}$\uparrow$ ($x_{t}$, $y_t$)} & \multicolumn{2}{c|}{\cellcolor{blue} $\uparrow$ ($x_{it}$, $x_t$)} & \multicolumn{2}{c|}{\cellcolor{red} $\downarrow$ ($x_{it}$, $x_i$)} & \multicolumn{2}{c}{ \cellcolor{blue} $\uparrow$($x_{it}$, $y_t$)}\tabularnewline
\cline{10-17} \cline{11-17} \cline{12-17} \cline{13-17} \cline{14-17} \cline{15-17} \cline{16-17} \cline{17-17} 
\multicolumn{6}{c}{} &  &  &  & {real} & {fake} & {real} & {fake} & {real} & {fake} & {real} & {fake}\tabularnewline
\cline{1-7} \cline{2-7} \cline{3-7} \cline{4-7} \cline{5-7} \cline{6-7} \cline{7-7} 
\cellcolor{red} $L_1$ & \cellcolor{red} $L_2$ &\cellcolor{blue} $L_3$& \cellcolor{blue} $L_4$ & \cellcolor{blue} $L_5$ & \cellcolor{red} $L_6$ & $\mathcal{R}(W)$ &  &  &  &  &  &  &  &  &  & \tabularnewline
\hline 
 &  &  &  &  &  &  & 56.72 & 33.04 & 76.27 & 61.88 & 98.87 & 95.64 & 89.97 & 89.53 & 62.57 & 48.52\tabularnewline
\hline
\hline
\checkmark &  &  &  &  &  & 0.5 & 41.30 & 34.01 & 2.11 & 0.08 & 7.78 & 1.46 & 99.02 & 99.19 & 0.15 & 0.03\tabularnewline
\hline 
 \checkmark &  &  &  &  & \checkmark & 0.5 & 49.92 & 40.96 & 5.87 & 0.3 & 13.51 & 2.81 & 98.34 & 98.88 & 0.38 & 0.04\tabularnewline
\hline 
 \checkmark & \checkmark &  &  &  & \checkmark & 0.5 & 51.52 & 41.39 & 8.47 & 0.5 & 21.21 & 4.96 & 97.57 & 98.28 & 0.57 & 0.04\tabularnewline
\hline 
 \checkmark & \checkmark &  &  & \checkmark & \checkmark & 0.5 & 50.37 & 40.62 & 1.39 & 0.09 & 9.14 & 1.98 & 97.84 & 98.42 & 0.18 & 0.05\tabularnewline
\hline 
 \checkmark & \checkmark & \checkmark &  & \checkmark & \checkmark & 0.5 & 49.68 & 40.05 & 0.08 & 0.00 & 10.67 & 2.8 & 98.01 & 98.56 & 0.13 & 0.04\tabularnewline
\hline 
\checkmark & \checkmark & \checkmark & \checkmark & \checkmark & \checkmark & 0.5 & 49.60 & 40.05 & 0.07 & 0.01 & 10.45 & 2.78 & 97.99 & 98.58 & 0.15 & 0.03\tabularnewline
\hline 
\hline
 \checkmark & \checkmark &  &  & \checkmark & \checkmark & 0.5 & 50.37 & 40.62 & 1.39 & 0.09 & 9.14 & 1.98 & 97.84 & 98.42 & 0.18 & 0.05\tabularnewline
\hline 
  \checkmark & \checkmark &  &  & \checkmark & \checkmark & 0.0 & 12.89 & 9.40 & 0.01 & 0.01 & 0.09 & 0.02 & 23.48 & 31.88 & 0.01 & 0.01\tabularnewline
\end{tabular}
}
\caption{The ablation of the effects of different loss terms across classification and retrieval tasks of the tuples on the validation set for the "forget to spell" model.}
    \label{tab:retireval_forget}
\end{table*}

\begin{figure*}
  \centering
  \includegraphics[width=\linewidth]{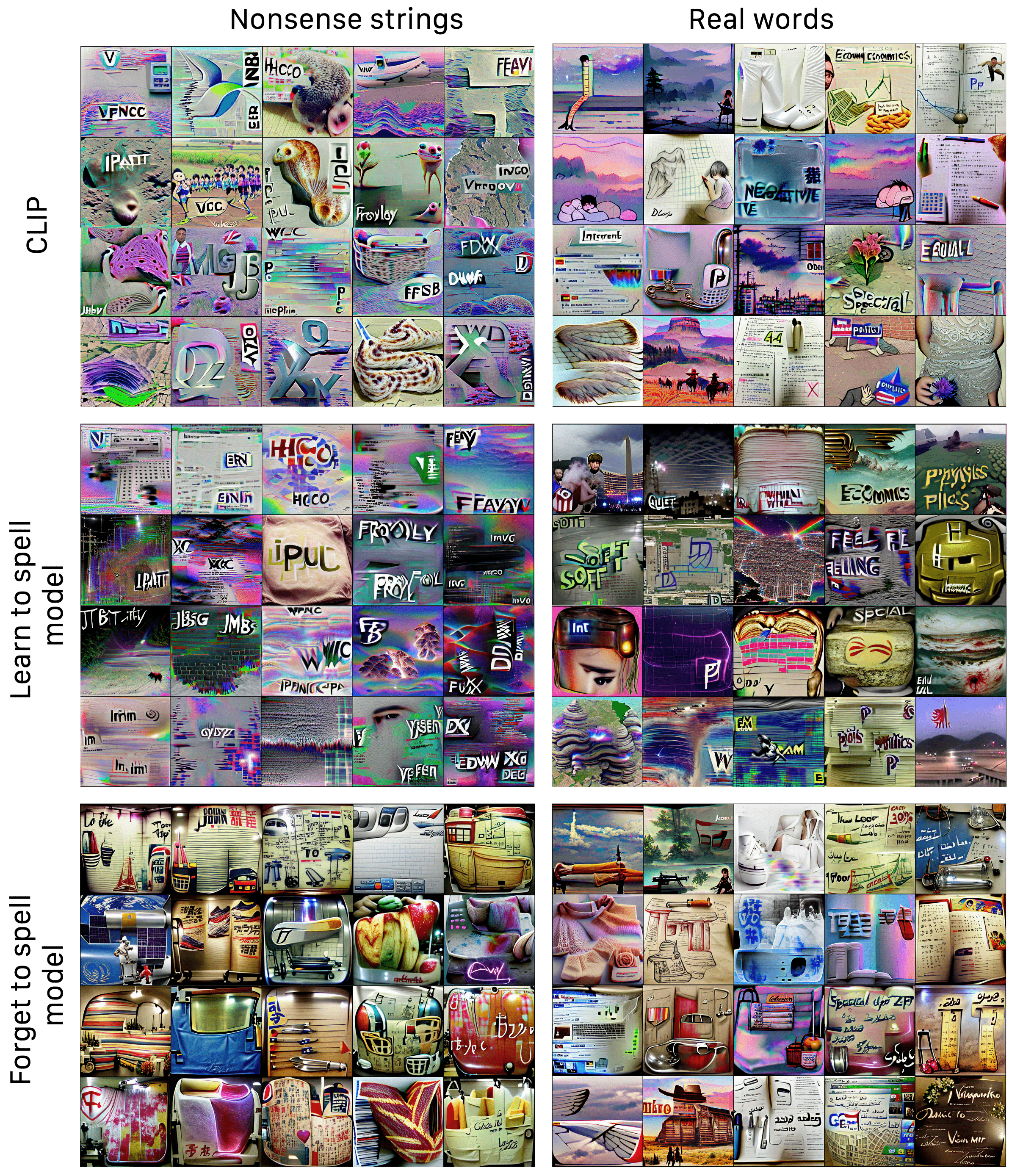}
  \caption{Images generated with text-conditioning using CLIP, "learn to spell" model, and "forget to spell" model. Text prompts used for nonsense strings (from left to right, starting from top left: 'vfnpcd', 'ebnr', 'hcioo', 'vhhh', 'feayv', 'jqtibdy', 'jlsbmg', 'wcpinc', 'fysllqb', 'duxwf', 'ipaut', 'vjcxc', 'ipcui', 'froyl', 'imcqvg', 'irmin', 'qzdyf', 'qhyx', 'yfeseni', 'xdegiw'. Text prompts used for real words: 'long', 'quiet', 'white', 'economics', 'physics', 'internet', 'private', 'ordinary', 'special', 'equal', 'soft', 'drawing', 'negative', 'feeling', 'homework', 'wing', 'western', 'exam', 'politics', 'formal'.
}
  \label{fig:big_generation}
\end{figure*}

For training the projection matrices, we take the ImageNet dataset, for each natural image and text class label $x_i, y_i$ we sample a string and generate a pair of a word image and a text string $x_t, y_t$, and a natural image with text $x_{it}$. The string $y_i$ is written as a text “an image of {class label}”. We use a corpus of 202587 English words, we use 182329 words in the training set and 20258 in the validation set, the words are all lower case, between 3 and 10 letters. For half of the tuples in our dataset we use nonsense strings, which are generated by uniformly sampling a length of the string (between 3 and 10), and sampling letters from the Latin alphabet. We are not using any prompts for the language embeddings and follow the image processing pipeline from \cite{radford2021learning}.

We train each projection matrix for 1 epoch, with learning rate 0.0001, step learning rate decay of factor 0.5 every 4000 steps with Adam optimizer. We use batch size 128. The size of the matrix $W$ is tuned for each task. For the “learn to spell” task, we test bottleneck dimensions between 32 and 512 with increment of 32, using only loss $L_4$ and $\gamma = 0.5$, the retrieval accuracy image to text on fake images is shown in Fig. \ref{fig:bottleneck}. The matrix with 512x512 dimensions achieves comparable performance to the original CLIP network, this is because the regularizer term forces the matrix W to be orthogonal, hence at the original dimension, we simply learn a rotation in the space, and the accuracy score remains (nearly) the same. We observe that the highest accuracy is reached at 64 dimensions, and steadily decreases when choosing a larger or smaller number. Intuitively, this suggests that the ability to recognize written text can be encoded in 64 dimensions. Our next ablations for this model are concerning a matrix 512x64 dimensions.  

We ablate different terms in of the $L_{spell}$ loss and report the results in Table~\ref{tab:retireval_learn}, for the tasks involving image classification we report top-1 accuracy, for the other tasks we report the retrieval score on the set of 20258 real words images and text and the same number of fake words for a fair comparison. We choose to report the score separately for the set and real and fake images, because the network has a prior knowledge about real words, and we want to test its generalization ability to any strings. The tasks that should improve are noted with $\uparrow$, and conversely the task that should impair are denoted with $\downarrow$. The columns marked blue are the ones corresponding to “learn to spell task”, we expect the performance of on those tasks to improve, and conversely the performance on the tasks marked with red to deteriorate. We can observe that the positive terms in the loss generally improve the performance of the model, albeit the full loss as show in \ref{eq:1} is not the best performing, as our final model we choose the model trained with $L_1, L_3, L_4, L_5$. We compare our best model with a model trained without the regularization term, we can see that it achieves lower performance by 10\% on the most important tasks involving correlating word images with text strings, and natural images with text with text strings (${(x_t,y_t), (x_{it}, y_t)}$).

Similarly, for the “forget to spell” model, we empirically find that the model performs the best at task 1 $(x_{it},x_i)$ with 256 dimensions. We present the ablations with different loss terms in Table~\ref{tab:retireval_forget}. We choose our final model as the model trained with combination of loss terms, $L_1, L_2, L_5, L_6$. In this case, we expect the performance of the tasks marked red to improve and the performance of the columns marked with blue to drop. Again, for this task, the orthogonality constraint is crucial.  We observe that the performance of the model trained without the orthogonal regularization term drops drastically for all the tasks. 

\begin{figure}
  \centering
  \includegraphics[width=0.9\linewidth]{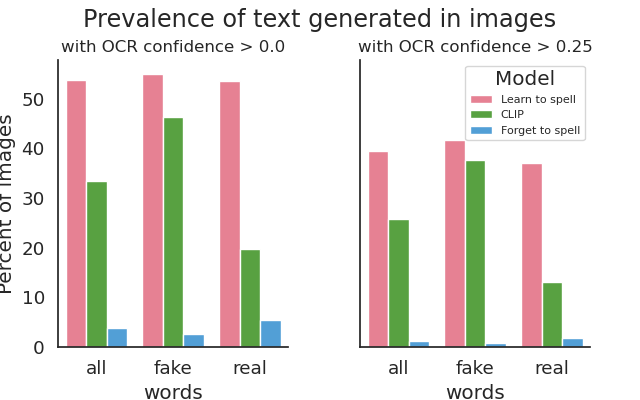}
   \caption{Text detection evaluation in images generated with different models.}
   \label{fig:prevalence}
\end{figure}

\begin{figure}
  \centering
  \includegraphics[width=0.85\linewidth]{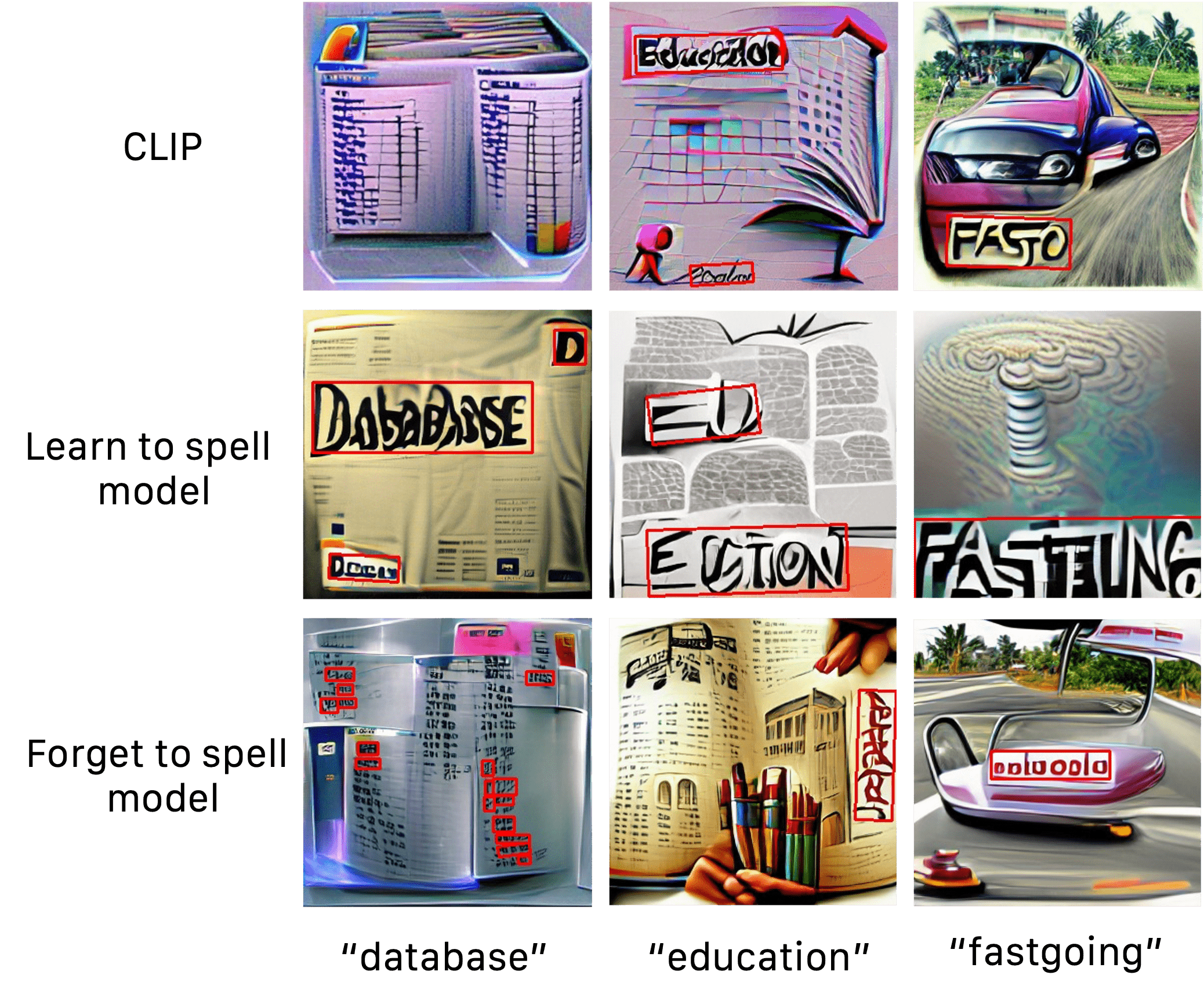}
   \caption{Qualitative examples of the OCR detection in the images generated using the CLIP model and our learned projections.}
   \label{fig:textdet}
\end{figure}

\section{Evaluation}
\subsection{Text Generation}
To visualize the written text (dis-)entanglement, we generate images conditioned on text prompts. We use an open-source implementation from \cite{crowsonvqgan} of a VQGAN generation model \cite{esser2021taming} which steers the image generation based on a text prompt. A discrete latent code is randomly sampled, and then optimized such that the cosine similarity between the CLIP embedding of a generated image and the CLIP embedding of the target text prompt is maximized.

To inspect our learned projections, we follow the same scheme, but compute the loss on the W-projections of the synthesized image and text CLIP embeddings. It is important to highlight that our goal is not a novel font synthesis or improving the quality of the text-to-image generation, but rather using this task as a lens into our learned projections. We generate 1000 images conditioned on real English words from our validation set, and 1000 images conditioned on nonsense strings from the validation text string set using VQGAN+CLIP and both of our projection models. Figure~\ref{fig:bigteaser} presents samples of generated images: the first row shows images generated with the original VQGAN+CLIP setting, capturing the visual concepts of the target prompts, and in cases of “peas”, “time”, “focus”, and “police” also showing the letters of the words. The “forget to spell” model is able to capture the visual concepts of the words without the letters, and the “learn to spell” model shows imperfect, but legible letters corresponding to the text prompt. Figure~\ref{fig:big_generation} shows more qualitative results, using both real and fake words as text prompts. In case of nonsense strings, the VQGAN+CLIP method is more likely to produce image text, possibly because nonsense string text prompts do not have a visual meaning associated with them. The images generated with the “forget to spell” model still contain text-like texture, but with less resemblance to the Latin alphabet than to Asian text forms.

To quantify the appearance of text, we detect words in images using an open-source OCR tool \cite{easyocr}. State-of-the art OCR recognition models are typically trained on either natural images with text \cite{gupta2016synthetic} or synthetic datasets of natural images with rendered text \cite{gupta2016synthetic}. While our generated images are much different from those training datasets, we qualitatively inspect the predictions and find them accurate (Figure \ref{fig:textdet}).
A text detection in an image is recognized if the area of the detected word is larger than 10\% of the area of the image and there are at least 2 letters in the predicted word that are the same as the target text prompt.

\begin{figure}
  \centering
  \includegraphics[width=0.85\linewidth]{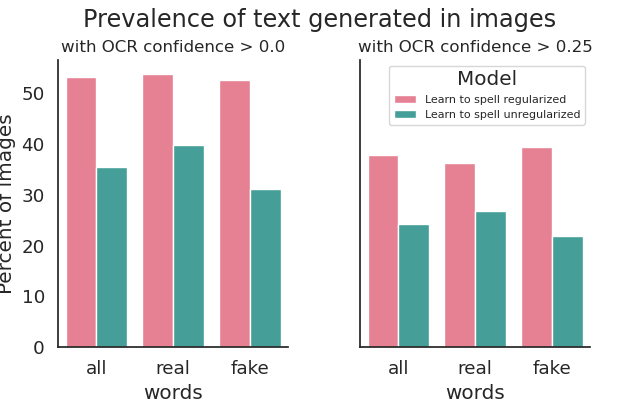}
   \caption{Word detection rates in "learn to spell" models trained with and without orthogonality constraint.}
   \label{fig:prevalence_unreg}
\end{figure}
\begin{figure}
  \centering
  \includegraphics[width=0.9\linewidth]{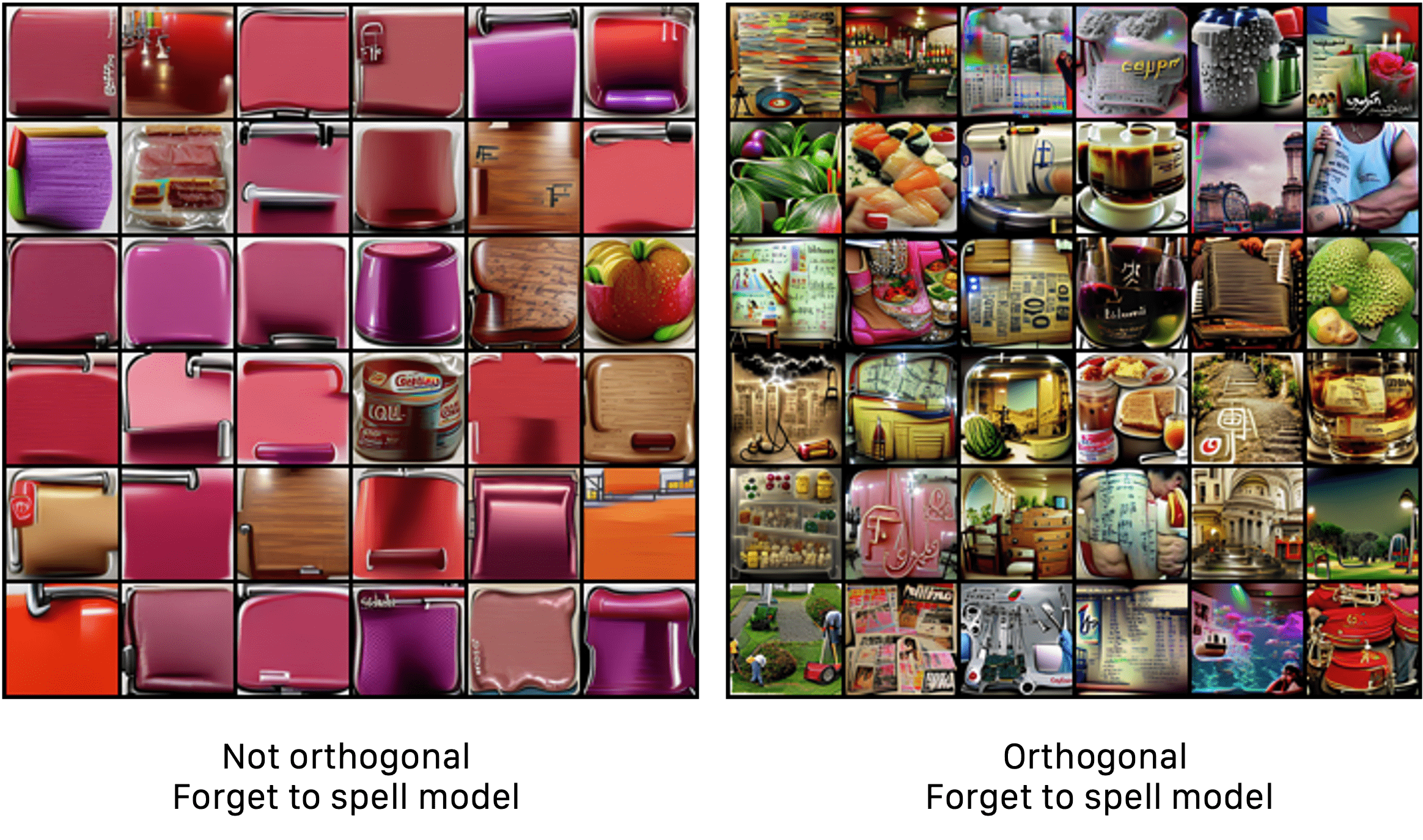}
   \caption{Images generated conditioned on regularized and  unregularized "forget to spell" model.}
   \label{fig:prevalence_unreg_image}
\end{figure}

Results of OCR text detection are shown in Figure~\ref{fig:prevalence}. The difference in all detections across all words between the original model and the “learn to spell” projection is 25.43\%, and between the “learn to spell” model and the “forget to spell” model is 54.92\%. The gap is more prominent when looking at real-word-conditioned generations, which confirms the qualitative analysis. The difference between the prevalence of detections is less significant in fake-word-conditioned generations, which we attribute to the fact that those words lack visual meaning.

\begin{figure*}[!ht]
  \centering
  \includegraphics[width=\textwidth]{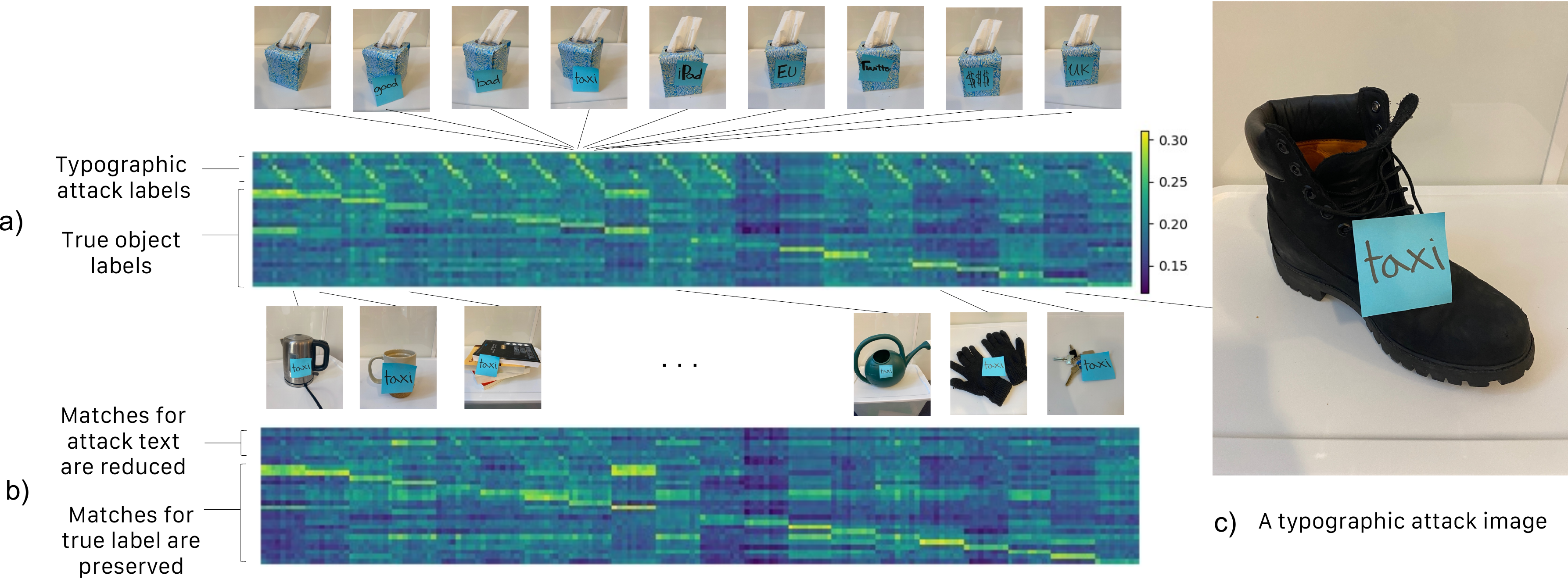}
  \caption{A test on a data set of 200 text attack images, a) shows a similarity matrix between the embeddings images with typographic attacks and the the text embeddings of typographic attack labels and true object labels obtained by the CLIP model, b) shows the same similarity matrix obtained by the Forget-to-Spell model.}
  \label{fig:apple_ipod}
\end{figure*}

\textbf{Non-orthogonal projections}
We compare the image generation experiments between the projections trained with and without orthogonal constraints. The orthogonal ``learn to spell'' model shows 17.5\% more text detections than its non-orthogonal comparison (Figure~\ref{fig:prevalence_unreg}). Similarly, we test the importance of orthogonality in the “forget to spell” model. While the detection rate in those images is close to 0\%, the images generated using non-orthogonal model have collapsed to a single pattern of red background (Figure~\ref{fig:prevalence_unreg_image}). Without the orthogonality constraint, the projection is no longer able to preserve the original CLIP model representations, and loses any meaning.

\subsection{Robustness}
Our second evaluation task is OCR. We consider the IIIT5K dataset \cite{mishra2012scene}, a dataset of natural images of cropped words.
We compute a retrieval score on the lexicon classification task (1 out of 1000), and a retrieval amongst all the unique words in the dataset (1 out of 1772). In the first task, our projection with 128 dimensions is able to achieve a performance only 1.76\% lower than the original 512-dimensional embedding, despite the testing task being out-of-domain. When testing on the full dataset, we see a 0.2\% improvement over the original CLIP model. When testing on a 64-dimensional projection, the orthogonal projection obtains a 4.87\% drop in performance, whereas the non-orthogonal projection suffers a 24.63\% drop (Table~\ref{tab:robustness_spell}).

To test the typographic attack setting, we collect a dataset of 180 images of 20 objects and 8 typographic attacks. The accuracy of CLIP on true object labels is only 49.4\%, whereas the “forget- to-spell” model obtains 77.2\%. Figure \ref{fig:apple_ipod} shows the full similarity matrices, in Figure \ref{fig:apple_ipod}a, the diagonal pattern for each object on all typographic attack labels shows that CLIP responds strongly to the text label, while in Figure \ref{fig:apple_ipod}b, this sensitivity to text is reduced. Sensitivity to the true object label is preserved. Note, that the projection matrices were trained to disentangle text in images only with synthetic text images, and the testing data shows natural images with text, which demonstrates the out-of-domain generalization of the Forget-to-spell model.

\begin{table}
  \centering
 \resizebox{\linewidth}{!}{
\begin{tabular}{c|c|c|c|c}
{Model} & {Dimension} & {Regularized} & \multicolumn{2}{c}{Accuracy}\tabularnewline
\cline{4-5} \cline{5-5} 
 &  &  & IIIT5K 1K & IIIT5K \tabularnewline
\hline 
\hline 
CLIP & 512 &  & 69.43 & 63.00\tabularnewline
\hline 
\hline 
Learn to spell & 128 & \checkmark & 67.67 & 63.20\tabularnewline
\hline 
Learn to spell & 128 &  &45.56 & 39.23\tabularnewline
\hline 
Learn to spell & 64 & \checkmark & 64.56 & 61.17 \tabularnewline
\hline 
Learn to spell & 64 &  & 44.80 & 39.00\tabularnewline
\end{tabular}}
\caption{Out-of-domain generalization evaluation on the IIIT5K dataset.}
    \label{tab:robustness_spell}
\end{table}

\section{Limitations}
\vspace{-0.05in}
Our method delivers orthogonal subspaces of the CLIP vector space that can generate images with more and fewer visual words in synthesized images. However, we can not perfectly avoid text all together when using the “forget to spell” projection, nor can we guarantee perfectly written text using the “learn to spell” projection. As seen in our qualitative (Figure~\ref{fig:big_generation}) and quantitative (Figure~\ref{fig:prevalence_unreg}) results, some target text prompts remain in generated images, and in others we can observe some letters from the target word. 
\vspace{-0.05in}
\section{Conclusion}
\vspace{-0.05in}
We have studied the relationship between rendered text and its visual meaning as represented by the CLIP network, motivating the problem with examples of text confusion when generating an image. We have found that a learned orthogonal projection is able to disentangle the written and visual comprehension in the CLIP image encoding; orthogonality is crucial for our method. We have explored two distinct applications: reducing text artifacts in text-to-image generation, and defense against typographic attacks, collecting an evaluation  dataset of typographic attack images to measure the latter. We find that our method is effective in both applications, controlling generation of text in images, and reducing text confusion in zero-shot classification.

{\section*{Acknowledgement}
\noindent
We are grateful to Manel Baradad for early feedback and valuable discussions. JM was partially funded by the MIT-IBM Watson AI Lab, and DB was supported by DARPA SAIL-ON HR0011-20-C-0022. 
}

\bibliographystyle{ieee_fullname}
\bibliography{egbib}

\begin{thebibliography}{10}\itemsep=-1pt

\bibitem{alain2016understanding}
Guillaume Alain and Yoshua Bengio.
\newblock Understanding intermediate layers using linear classifier probes.
\newblock In {\em ICLR Workshop}, 2016.

\bibitem{bau2021paint}
David Bau, Alex Andonian, Audrey Cui, YeonHwan Park, Ali Jahanian, Aude Oliva,
  and Antonio Torralba.
\newblock Paint by word.
\newblock {\em arXiv preprint arXiv:2103.10951}, 2021.

\bibitem{collins2020editing}
Edo Collins, Raja Bala, Bob Price, and Sabine Susstrunk.
\newblock Editing in style: Uncovering the local semantics of gans.
\newblock In {\em Proceedings of the IEEE/CVF Conference on Computer Vision and
  Pattern Recognition}, pages 5771--5780, 2020.

\bibitem{crowson2021vqganclip}
Katherine Crowson.
\newblock {VQGAN+CLIP}.
\newblock
  \sloppy{\url{https://colab.research.google.com/drive/15UwYDsnNeldJFHJ9NdgYBYeo6xPmSelP}},
  Jan. 2021.

\bibitem{crowsonvqgan}
Katherine Crowson.
\newblock {VQGAN+pooling}.
\newblock \sloppy{\url{
  https://colab.research.google.com/drive/1ZAus_gn2RhTZWzOWUpPERNC0Q8OhZRTZ}},
  Jan. 2021.

\bibitem{esser2021taming}
Patrick Esser, Robin Rombach, and Bjorn Ommer.
\newblock Taming transformers for high-resolution image synthesis.
\newblock In {\em Proceedings of the IEEE/CVF Conference on Computer Vision and
  Pattern Recognition}, pages 12873--12883, 2021.

\bibitem{fong2018net2vec}
Ruth Fong and Andrea Vedaldi.
\newblock Net2vec: Quantifying and explaining how concepts are encoded by
  filters in deep neural networks.
\newblock In {\em Proceedings of the IEEE conference on computer vision and
  pattern recognition}, pages 8730--8738, 2018.

\bibitem{goetschalckx2019ganalyze}
Lore Goetschalckx, Alex Andonian, Aude Oliva, and Phillip Isola.
\newblock Ganalyze: Toward visual definitions of cognitive image properties.
\newblock In {\em CVPR}, pages 5744--5753, 2019.

\bibitem{goh2021multimodal}
Gabriel Goh, Nick Cammarata, Chelsea Voss, Shan Carter, Michael Petrov, Ludwig
  Schubert, Alec Radford, and Chris Olah.
\newblock Multimodal neurons in artificial neural networks.
\newblock {\em Distill}, 6(3):e30, 2021.

\bibitem{gupta2016synthetic}
Ankush Gupta, Andrea Vedaldi, and Andrew Zisserman.
\newblock Synthetic data for text localisation in natural images.
\newblock In {\em Proceedings of the IEEE conference on computer vision and
  pattern recognition}, pages 2315--2324, 2016.

\bibitem{harkonen2020ganspace}
Erik H{\"a}rk{\"o}nen, Aaron Hertzmann, Jaakko Lehtinen, and Sylvain Paris.
\newblock Ganspace: Discovering interpretable gan controls.
\newblock {\em arXiv preprint arXiv:2004.02546}, 2020.

\bibitem{jahanian2020steerability}
Ali Jahanian, Lucy Chai, and Phillip Isola.
\newblock On the "steerability" of generative adversarial networks.
\newblock In {\em ICLR}, 2020.

\bibitem{easyocr}
JaidedAI.
\newblock {EasyOCR}.
\newblock \sloppy{{\url{https://github.com/JaidedAI/EasyOCR}}},, 2021.

\bibitem{karras2020analyzing}
Tero Karras, Samuli Laine, Miika Aittala, Janne Hellsten, Jaakko Lehtinen, and
  Timo Aila.
\newblock Analyzing and improving the image quality of stylegan.
\newblock In {\em Proceedings of the IEEE/CVF Conference on Computer Vision and
  Pattern Recognition}, pages 8110--8119, 2020.

\bibitem{kim2018interpretability}
Been Kim, Martin Wattenberg, Justin Gilmer, Carrie Cai, James Wexler, Fernanda
  Viegas, et~al.
\newblock Interpretability beyond feature attribution: Quantitative testing
  with concept activation vectors (tcav).
\newblock In {\em International conference on machine learning}, pages
  2668--2677. PMLR, 2018.

\bibitem{lemesle2022language}
Yoann Lemesle, Masataka Sawayama, Guillermo Valle-Perez, Maxime Adolphe,
  H{\'e}l{\`e}ne Sauz{\'e}on, and Pierre-Yves Oudeyer.
\newblock Language-biased image classification: Evaluation based on semantic
  compositionality.
\newblock In {\em International Conference on Learning Representations}, 2022.

\bibitem{mishra2012scene}
Anand Mishra, Karteek Alahari, and CV Jawahar.
\newblock Scene text recognition using higher order language priors.
\newblock In {\em BMVC-British Machine Vision Conference}. BMVA, 2012.

\bibitem{murdock2021bigsleep}
Ryan Murdock.
\newblock {The Big Sleep}.
\newblock \sloppy{\url{
  https://colab.research.google.com/drive/1NCceX2mbiKOSlAd_o7IU7nA9UskKN5WR}},
  Jan. 2021.

\bibitem{patashnik2021styleclip}
Or Patashnik, Zongze Wu, Eli Shechtman, Daniel Cohen-Or, and Dani Lischinski.
\newblock Styleclip: Text-driven manipulation of stylegan imagery.
\newblock In {\em Proceedings of the IEEE/CVF International Conference on
  Computer Vision}, pages 2085--2094, 2021.

\bibitem{radford2021learning}
Alec Radford, Jong~Wook Kim, Chris Hallacy, Aditya Ramesh, Gabriel Goh,
  Sandhini Agarwal, Girish Sastry, Amanda Askell, Pamela Mishkin, Jack Clark,
  et~al.
\newblock Learning transferable visual models from natural language
  supervision.
\newblock {\em arXiv preprint arXiv:2103.00020}, 2021.

\bibitem{ramesh2021zero}
Aditya Ramesh, Mikhail Pavlov, Gabriel Goh, Scott Gray, Chelsea Voss, Alec
  Radford, Mark Chen, and Ilya Sutskever.
\newblock Zero-shot text-to-image generation.
\newblock {\em arXiv preprint arXiv:2102.12092}, 2021.

\bibitem{russakovsky2015imagenet}
Olga Russakovsky, Jia Deng, Hao Su, Jonathan Krause, Sanjeev Satheesh, Sean Ma,
  Zhiheng Huang, Andrej Karpathy, Aditya Khosla, Michael Bernstein, et~al.
\newblock Imagenet large scale visual recognition challenge.
\newblock {\em International journal of computer vision}, 115(3):211--252,
  2015.

\bibitem{shen2020interfacegan}
Yujun Shen, Ceyuan Yang, Xiaoou Tang, and Bolei Zhou.
\newblock Interfacegan: Interpreting the disentangled face representation
  learned by gans.
\newblock {\em IEEE transactions on pattern analysis and machine intelligence},
  2020.

\bibitem{wu2021stylespace}
Zongze Wu, Dani Lischinski, and Eli Shechtman.
\newblock Stylespace analysis: Disentangled controls for stylegan image
  generation.
\newblock In {\em Proceedings of the IEEE/CVF Conference on Computer Vision and
  Pattern Recognition}, pages 12863--12872, 2021.

\bibitem{zhou2017places}
Bolei Zhou, Agata Lapedriza, Aditya Khosla, Aude Oliva, and Antonio Torralba.
\newblock Places: A 10 million image database for scene recognition.
\newblock {\em IEEE Transactions on Pattern Analysis and Machine Intelligence},
  2017.

\bibitem{zhou2018interpretable}
Bolei Zhou, Yiyou Sun, David Bau, and Antonio Torralba.
\newblock Interpretable basis decomposition for visual explanation.
\newblock In {\em Proceedings of the European Conference on Computer Vision
  (ECCV)}, pages 119--134, 2018.

\end{thebibliography}

\end{document}